\documentclass[letterpaper, 10 pt, conference]{ieeeconf}  

\IEEEoverridecommandlockouts                              

\usepackage{graphicx}
\usepackage{amssymb}
\graphicspath{{Figure/}}
\usepackage{amsmath}
\usepackage[table,xcdraw]{xcolor}
\usepackage{booktabs}

\usepackage{subcaption}	
\usepackage[colorlinks,linkcolor=red,anchorcolor=blue,citecolor=blue]{hyperref}
\usepackage{cleveref}

\title{\LARGE \bf
Co-Fix3D: Enhancing 3D Object Detection With Collaborative Refinement
}

\author{Wenxuan Li, Qin Zou, Chi Chen, Bo Du, Long Chen, Jian Zhou, Hongkai Yu
\thanks{W. Li, Q. Zou and B. Du are with the School of Computer Science, Wuhan University, Wuhan 430072, China. (Emails: \{liwenx, qzou, dubo\}@whu.edu.cn)}
\thanks{C.~Chen  and  Jian Zhou are with the State Key Laboratory of Surveying, Mapping, and Remote Sensing Information Engineering, Wuhan University, Wuhan 430079, China (e-mail:\{chichen, jianzhou\} @whu.edu.cn).}
\thanks{L. Chen is with the Institute of Automation, Chinese Academy of Sciences, Beijing 130028, China. (Email: long.chen@ia.ac.cn)}
\thanks{H. Yu is with the Department of Electrical and Computer Engineering, Cleveland State University, Cleveland, OH 44115.}
}

\begin{document}

\maketitle
\thispagestyle{empty}
\pagestyle{empty}

\begin{abstract}
		\noindent 
		3D object detection in driving scenarios faces the challenge of complex road environments, which can lead to the loss or incompleteness of key features, thereby affecting perception performance. To address this issue, we propose an advanced detection framework called Co-Fix3D. Co-Fix3D integrates Local and Global Enhancement (LGE) modules to refine Bird's Eye View (BEV) features. The LGE module uses Discrete Wavelet Transform (DWT) for pixel-level local optimization and incorporates an attention mechanism for global optimization. To handle varying detection difficulties, we adopt multi-head LGE modules, enabling each module to focus on targets with different levels of detection complexity, thus further enhancing overall perception capability. Experimental results show that on the nuScenes dataset's LiDAR benchmark, Co-Fix3D achieves 69.4\% mAP and 73.5\% NDS, while on the multimodal benchmark, it achieves 72.3\% mAP and 74.7\% NDS. The source code is publicly available at \href{https://github.com/rubbish001/Co-Fix3d}{https://github.com/rubbish001/Co-Fix3d}.
		
	\end{abstract}
	
	\section{INTRODUCTION}
3D object detection\cite{hu2023aerial} is essential for autonomous vehicles and robotics, facilitating precise identification and localization of objects in their environments. This field has advanced significantly with sophisticated 3D neural network models such as Convolutional Neural Networks (CNNs)\cite{feng2023clustering} and transformer technologies\cite{liu2023petrv2}. However, challenges in 3D object detection persist as images lack depth information for indicating 3D positions, and point clouds, while providing spatial details, often lack rich semantic information and struggle to capture distant objects due to sparsity. These characteristics present significant obstacles in the field.
	\begin{figure}[t]
		\centering
		\includegraphics[width=1.0\linewidth]{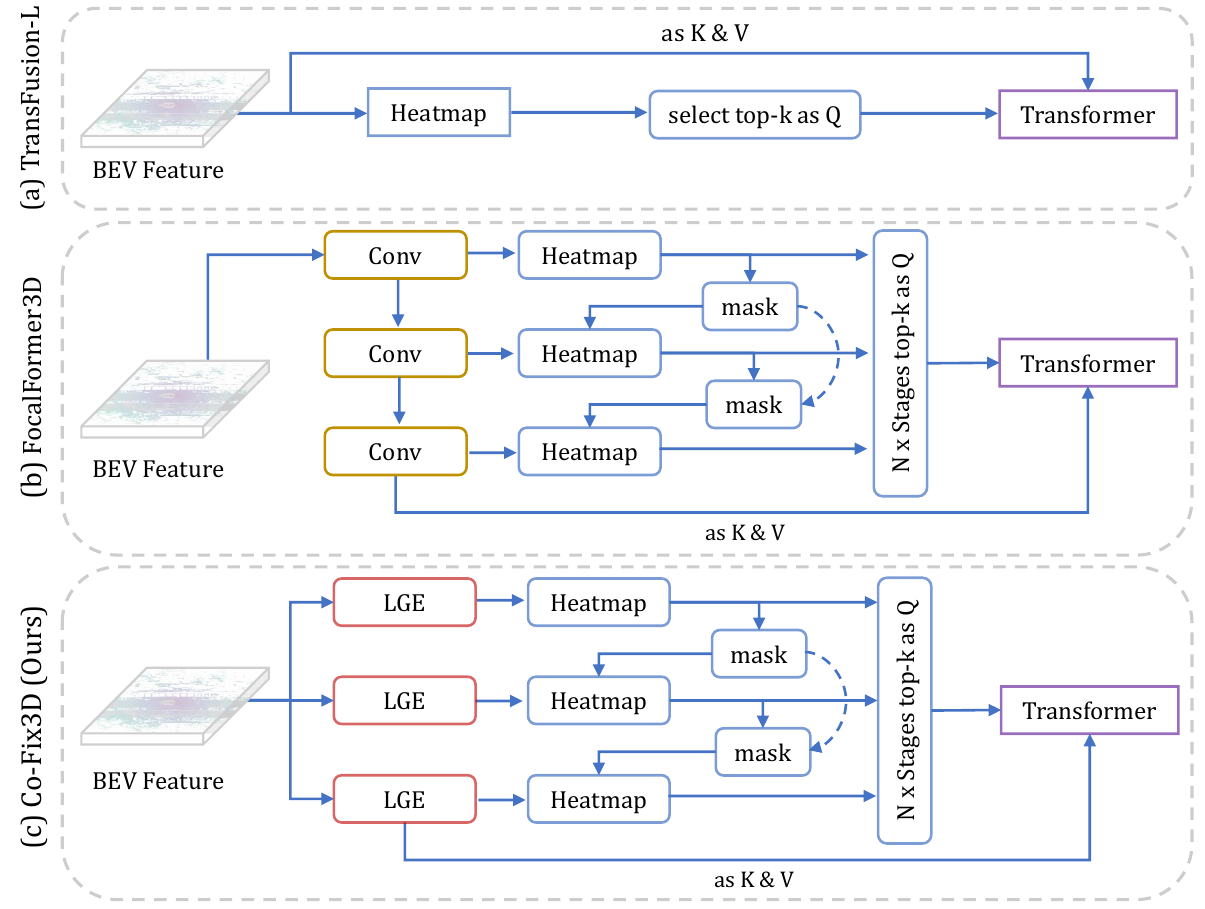}	
		\caption{
		Comparing TransFusion, FocalFormer3D, and our proposed Co-Fix3D.
		(a) TransFusion utilizes BEV features to generate heatmap score maps, selecting the highest-scoring K cells as queries. The quality of these queries directly impacts TransFusion's performance.
		(b)	FocalFormer3D uses a multi-stage approach with masking techniques to filter out easily detectable targets early. This increases queries and boosts detection by improving recall rates.
		(c) Co-Fix3D enhances BEV features with the LGE module and uses a parallel structure to adaptively process sample features across stages. This method generates high-quality query sets for accurate predictions in the Transformer module. }
		\label{dif}
		\vspace{-2mm}
	\end{figure}
	
To address these challenges, modern 3D object detection systems increasingly leverage BEV representations. This method provides a clear and effective spatial layout of objects, significantly enhancing the system's operational efficiency and decision-making capabilities. For example, TransFusion\cite{bai2022transfusion} (see Fig.\ref{dif}.(a)) merges BEV with transformer technology. It breaks down the 3D object detection process into two phases: initial coarse predictions using heatmaps, followed by refinements with transformer technology for improved accuracy. Additionally, FocalFormer3D\cite{chen2023focalformer3d} (see Fig.\ref{dif}.(b)) adopts a multi-stage approach to filter potential targets, effectively enlarging the candidate pool and boosting the recall rate to enhance detection performance.
However, BEV 3D object detection algorithms often overlook the uneven spatial distribution of point cloud data, which is denser near the sensor and sparser at distance. This oversight leads to inaccurate representation of features for distant targets. Moreover, the absence of depth information in image data leads to distorted and abnormal image BEV features. These issues significantly challenge accurate identification of hard samples in scenarios with occlusions, long distances, small targets, or complex backgrounds. To address these flawed BEV features, effectively extracting key information from BEV features and optimizing BEV features may be an effective solution.

In this study, we introduce Co-Fix3D (see Fig. \ref{dif}.(c)), an advanced detection framework designed specifically to address the complex challenges of 3D object detection. Inspired by the successful application of Discrete Wavelet Transform (DWT) in image restoration and its capability to decompose image data across different scales and capture multi-level features, we have employed DWT to optimize point cloud processing. DWT effectively separates high-frequency components (representing noise or details) from low-frequency components (representing smooth trends), which not only helps in noise reduction but also preserves crucial structural features, thereby enhancing the quality and usability of point cloud data. However, since DWT is primarily focused on pixel-level optimization, to further enhance the model's perception of the global context, we have incorporated an attention mechanism. Consequently, we developed a Local-Global Enhancement (LGE) module that performs adaptive cross-stage denoising and feature enhancement, significantly improving detection performance, especially by enhancing the identification and scoring of weak positive samples.
	
Furthermore, inspired by the multi-stage filtering mechanism of 3D detectors \cite{chen2023focalformer3d}, our experiments revealed that a parallel LGE architecture significantly enhances perception capabilities, in stark contrast to traditional cascading methods. Traditional cascading methods may impair the characteristics of undetected samples in the initial stages, making them harder to detect in subsequent stages. Additionally, we found that increasing the number of queries during the testing phase allows the parallel LGE detector to further enhance its perceptual abilities, an improvement not achieved with conventional cascading LGE methods. Overall, our approach significantly enhances the detection accuracy of small and partially obscured objects in complex environments, providing new insights into addressing the ongoing challenges of 3D object detection.

	In summary, our contribution can be summarized as follow:
	
	 i) Co-Fix3D, a multi-stage parallel architecture 3D detection network, is proposed to enhance BEV features end-to-end, enabling the precise identification of challenging instances.
	 
	ii) The LGE module is introduced to optimize BEV features, significantly enhancing the detection of weak positive samples.
	
	 iii) New benchmarks on the nuScenes 3D detection leaderboard have been established by our model, advancing the research in this field.
	
	\section{Related Work}
	\subsection{LiDAR-based 3D Object Detection.}
	
	LIDAR-based 3D object detection technologies\cite{song2023psns,luo2024gatr} are primarily categorized into three types: Point-based, Voxel-based, and Hybrid approaches. Point-based methods, such as PointNet\cite{qi2017pointnet} and PointNet++\cite{qi2017pointnet++}, directly process raw LiDAR data to extract critical features, enabling precise segmentation and refinement in models like PointRCNN \cite{shi2019pointrcnn}and VoteNet\cite{qi2019deep}. Voxel-based methods, including VoxelNet\cite{zhou2018voxelnet} and SECOND\cite{yan2018second}, transform point clouds into structured grids, facilitating efficient feature extraction while preserving accuracy. 
	Hybrid methods like PV-RCNN\cite{bhattacharyya2020deformable} combine the strengths of point-based and voxel-based techniques to enhance precision and efficiency. 
	
Currently, dense BEV detection technologies like TransFusion generally outperform sparse detectors in point cloud processing. Their successor, FocalFormer3D~\cite{chen2023focalformer3d}, improves detection by increasing the number of queries, which boosts the chances of selecting positive samples. However, when dealing with low reflectance, or distant objects in real scenarios, these 3D detectors often miss BEV feature defects that could degrade detection results. To tackle this, we introduced the LGE module. This module repairs weak BEV features and enhances their scores during the encoding stage, increasing positive sample queries and ensuring precise 3D object detection.
	
	\begin{figure*}[t]
		\centering
		\includegraphics[width=1.0\linewidth]{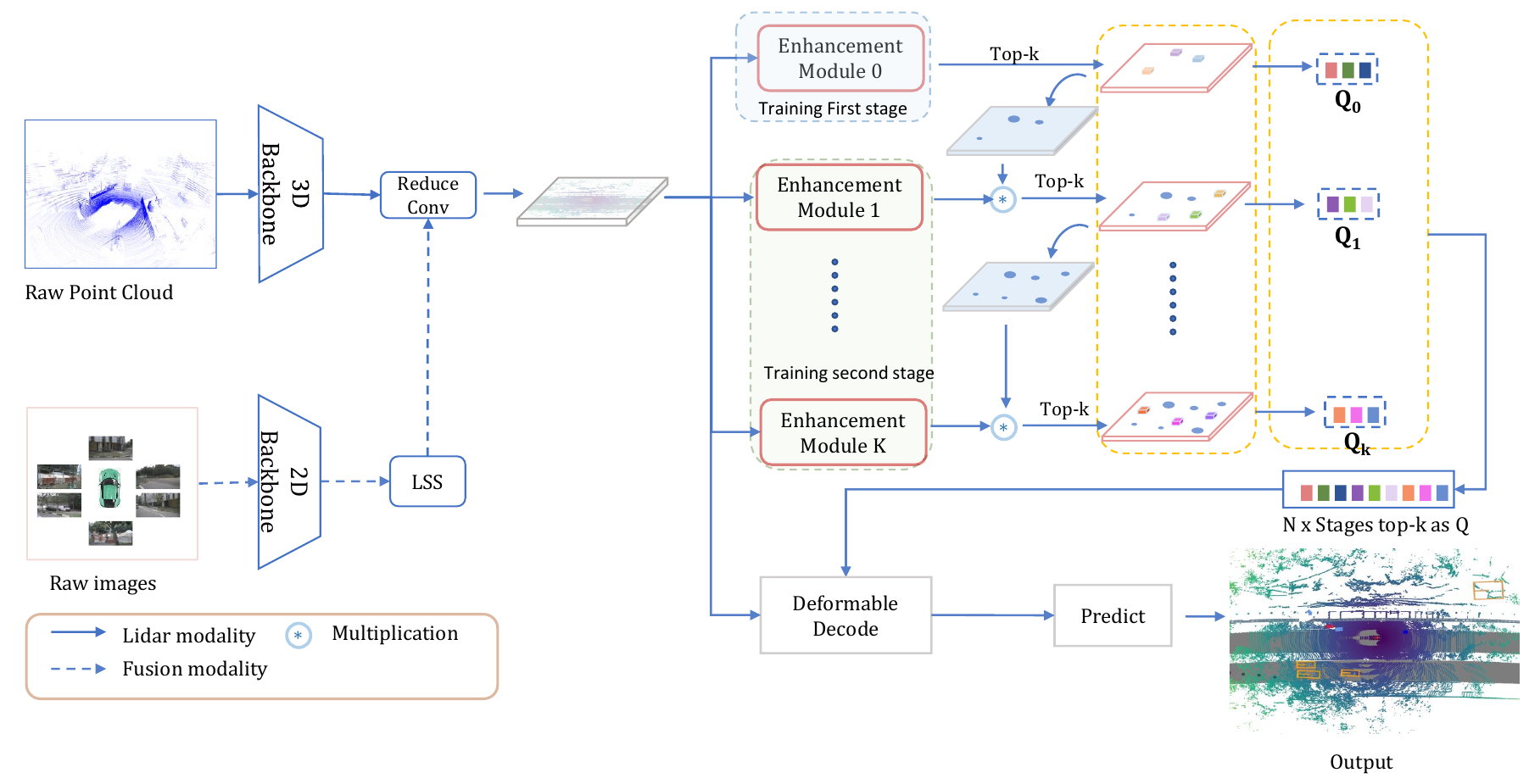}	
		\caption{			
			Overview of Co-Fix3D. Raw point cloud data is processed through a 3D Backbone network to generate LiDAR BEV features, while image data is processed through a 2D network and LSS to produce image BEV features. These features are fused into a new BEV representation using a Reduce Conv module and subsequently optimized through a multi-stage process leveraging Enhancement Module (LGE). At each stage, a top-k strategy is employed to select the highest-scoring queries for that stage, with a mask applied to prevent overlap between the selected queries and those from the previous stage. Finally, the $K\times N$ candidates are decoded to produce the detection outputs.}
		
		\label{main}
		\vspace*{-5mm}
	\end{figure*}
	\subsection{LiDAR-camera Fusion for 3D Object Detection.}
	LiDAR-camera fusion for 3D object detection\cite{chen2017multi} has become increasingly significant, with multimodal approaches often outperforming unimodal learning in capturing accurate latent space representations. These fusion methods can be categorized into early, middle, and late stages based on the timing of data integration.
	Early fusion methods\cite{chen2022deformable,xu2021fusionpainting}, exemplified by pioneering works like  enhance input points with corresponding image pixel features. However, they are sensitive to calibration errors. Late fusion approaches \cite{bai2022transfusion}, such as those in , fuse multimodal information at the region proposal level, often resulting in limited interactions between modalities and suboptimal detection performance. In contrast, middle fusion\cite{li2022voxel}, increasingly popular, promotes multimodal feature interaction at various stages, making it more robust to calibration errors. 
	
Building on this foundation, our proposed method, Co-Fix3D, employs an intermediate fusion strategy that integrates image data into BEV features using the Lift-Splat-Shoot (LSS)\cite{philion2020lift} method. However, these BEV representations often contain flaws that can lead to suboptimal detection outcomes. To improve this, Co-Fix3D utilizes the LGE module to enhance these features, significantly increasing their effectiveness and ensuring robust performance under challenging conditions.
	
	\section{Method}
	\subsection{Overview} 	

The overall workflow of Co-Fix3D is illustrated in Figure~\ref{main}. Co-Fix3D supports two detection modes: single-modal detection based solely on point clouds, and multi-modal detection that integrates both point cloud and camera data. In both modes, inputs are processed by an encoder to generate BEV features, which are then refined through a multi-stage enhancement module (LGE). During each stage, the LGE module uses a masking filter mechanism to select the top 
$k$ cells as queries. These selected cells are then processed by a deformable decoder for final object classification and localization.

Training occurs in two phases. In the first phase, only the initial enhancement module $0$ and the deformable decoder are trained. In the second phase, the remaining enhancement modules 
$1,2,…,K$ and $0$ are trained jointly with the deformable decoder. Subsequent enhancement modules progressively detect objects missed in previous stages, thus incrementally improving detection capabilities.
	
	\noindent{\textbf{Input Encoding.}} For single-modality detection, after processing through a 3D backbone network and associated flattening operations, we obtain the point cloud's BEV features, denoted as $F_{LiDAR} \in \mathbb{R}^{H \times W \times 4C}$, where $W$,$H$, and $C$ represent the width, height, and number of channels of the BEV feature map, respectively. Similarly, for the multimodal mode, after processing through a 2D backbone network and applying the original LSS method (without depth loss computation) \cite{philion2020lift}, we obtain the image's BEV features, denoted as $F_{Camera} \in \mathbb{R}^{H \times W \times C}$.  In the single-modality, $Reduce Conv$ module reduces the number of channels from	$4C$ to $C$; in the fusion mode, it reduces from $4C + C$ to $C$, ultimately forming the initial BEV feature $F_0$.

	\noindent{\textbf{Multi-stage Feature Enhancement.}} We use a multi-stage approach to generate queries, employing a mask mechanism to filter each stage progressively, allowing these parallel LGE modules to supervise different ground truths.	The BEV features $F_0$ are optimized within the LGE module and used to generate corresponding BEV heatmaps $H \in \mathbb{R}^{H \times W \times c}$,where $c$ represent the  category. 
	 We fistly initialized a mask $M \in {0, 1}^{H \times W \times 1}$, set entirely to 1. For the $(w, h)$ position and category $c$ of the heatmap at stage $i$, we used Top-k selection on the heatmap to set $k$ instances of $M_{i}(w,h,c)$ to 0. This indicates that once a region is selected, subsequent stages will not re-explore that region. We then applied box-level pooling methods to handle these 0-marked masks, ensuring that the generated query locations are as evenly distributed within the BEV as possible.
	To ensure diversity in the samples processed by each module after introducing the LGE module, we multiply the mask by the GT heatmap. This guarantees that different LGE modules repair different targets, enhancing the perception of targets with varying degrees of damage and maximizing the potential for target recognition. Specifically, if early-stage LGE modules fail to detect certain samples, subsequent stages will continue to monitor and learn from these samples until the targets are detected or the maximum number of stages is reached. 
	
	\subsection{Local and Global Enhancement Module}
	The LGE module is designed to eliminate noise and correct distorted features in BEV features. This module effectively integrates local and global denoising methods to enhance the accuracy and efficiency of data processing. It consists of three main components: the Wavelet Encode module for local optimization, the Hybrid Encode module for global optimization, and the Wavelet Decode module for post-processing. Next, we will first explain these three modules in detail one by one, and then introduce the various attempts made during the design of the LGE.
	
	\begin{figure}[h]
		\centering
		\includegraphics[width=0.9\linewidth]{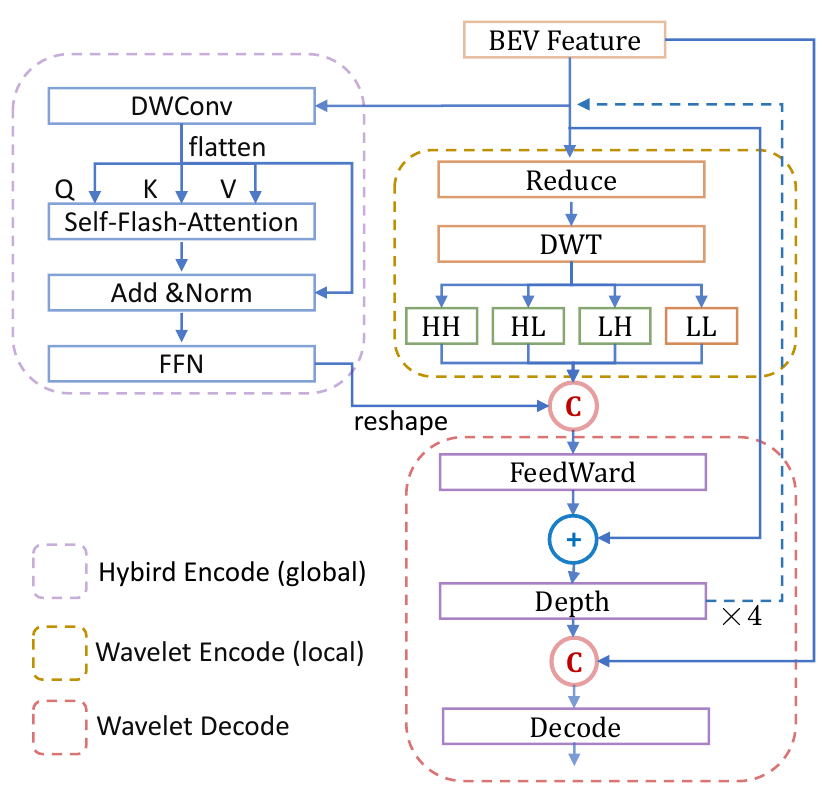}	
		\caption{Details of the LGE Module.}
		\label{detail_lge}
		\vspace*{-2mm}
	\end{figure}
	
	\noindent{\textbf{Wavelet Encode.}} Following the significant success of DWT in image restoration and super-resolution\cite{chen2024low,ji2023ultra}, we also leverage wavelet encoding using DWT to effectively restore the features of BEV grids, as shown in Figure \ref{detail_lge}. DWT compresses data by focusing on significant wavelet components and removing redundant information, effectively isolating and mitigating noise and anomalies in BEV features. This capability is particularly useful for restoring BEV features as it efficiently handles large-scale point cloud datasets.
	DWT decomposes BEV features into four distinct channels: HH, HL, LH, and LL, each capturing specific information aspect. The specific calculation process is as follows:
	\begin{align} \label{eq:1}
		& 	{F}_{1}={Reduce}({F_0)},\\
		& {F}_{LL},{F}_{LH},{F}_{HL},{F}_{HH}= DWT({F}_{1} ),\\
		&{F}_{2} = Concat({F}_{LL},{F}_{LH}, {F}_{HL},{F}_{HH}),
	\end{align}
	where $Reduce(\cdot)$ refers to reducing the number of channels of \( F_0 \) from \( C \) to \( \frac{C}{4} \), resulting in \( F_1 \in \mathbb{R}^{H \times W \times \frac{C}{4}} \). After applying $DWT(\cdot)$, \( F_{LL}, F_{LH}, F_{HL}, F_{HH} \in \mathbb{R}^{\frac{H}{2} \times \frac{W}{2} \times \frac{C}{4}} \). Finally, Concat concatenates these $DWT(\cdot)$ results along the channel dimension, leading to $ F_{2} \in \mathbb{R}^{\frac{H}{2} \times \frac{W}{2} \times C}$.
	
	\noindent{\textbf{Hybrid Encode.}}
	$Hybrid\  Encode$ (see Fig.~\ref{detail_lge}) employs a global attention mechanism to capture comprehensive global contextual information from BEV features, effectively minimizing noise and artifacts. This enhancement allows for a clearer and more precise distinction of complex sample features. Additionally, it integrates Flash Attention V2 \cite{dao2023flashattention}, greatly improving the efficiency of attention computations. In this module, BEV features $F_0$ are processed through a down-sampling layer and then flattened for the self-attention phase to assess feature importance. The process is described as follows:
	\begin{align} 
		& S_1=DWConv(F_{0}), \\
		&Q =K=V =Flatten(S_1), \\  
		&Q = Attn(Q, K, V),\\  
		&F_3 = Reshape(FFN(Q)),
	\end{align}
	where $DWConv(\cdot)$ denotes down-sampling, resulting in $S_{1} \in \mathbb{R}^{\frac{H}{2} \times \frac{W}{2} \times C}$.  Here, $Attn(\cdot)$ represents the multi-head self-attention mechanism. Finally, after processing through the feed-forward network ($FFN(\cdot)$), the output is reshaped using $Reshape(\cdot)$ to match the feature dimensions of $F_2$.
	
	\noindent{\textbf{{Wavelet Decode.}}
		The $Wavelet\ Decode$ module(see Fig.\ref{detail_lge} ) conducts post-processing. It primarily functions as a feedforward neural network and restores the resolution of BEV features. The process can be outlined as follows:
		\begin{align} 
			&S_2= FW(Concat(F_2,F_3)),\\
			&F_4 = Depth(F_p + S_2)),\\
			&F_5=	Decode(Concat(F_4,F_0)),
		\end{align}
		where $FW(\cdot)$ denotes a feedforward  wavelet network which performs up-sampling to restore the original shape,  resulting in $S_2 \in \mathbb{R}^{H \times W \times C}$. $Depth(\cdot)$ refers to an intermediate neural network that expands the depth of the network. Finally, $Decode(\cdot)$ simplifies the channel count, compressing the data for subsequent processing. 
		
		\subsection{Design choices of LGE}
		To design an optimized LGE structure, we conducted multiple attempts as shown in Figure \ref{design_lge}. Next, different forms of encoder are inserted to produce a
		series of variants, elaborated as follows:
		\begin{figure}[t]
			\centering
			\includegraphics[width=0.9\linewidth]{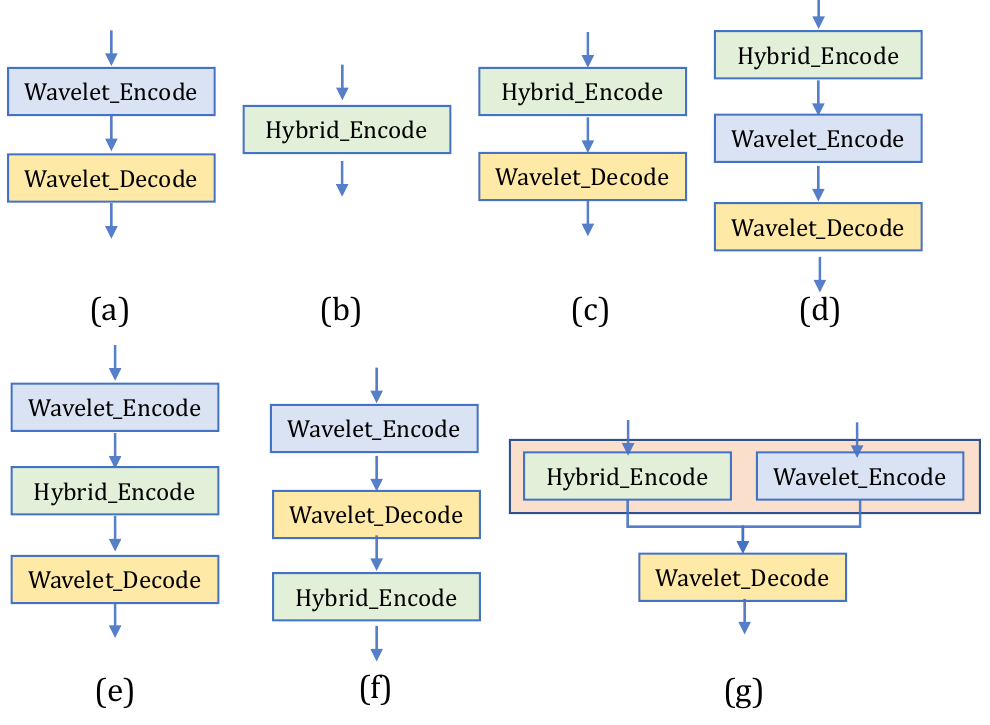}	
			\caption{The set of variants with different types of encoders.}
			\label{design_lge}
			\vspace*{-2mm}
		\end{figure}
		\begin{table*}[t]
	\centering
	\renewcommand{\arraystretch}{1.0}
	
	\begin{center}
			\caption{
			{Comparison with SOTA detectors on  nuScenes \textbf{TEST} set.} 
			We do not use test-time augmentation or model ensemble. 
			Mod.: Modality.
			C.V.: construction vehicle.
			Ped.: pedestrian.
			Mot.: motorcyclist.
			Byc.: bicyclist.
			T.C.: traffic cone.
			Bar.: barrier.
		}
		\begin{tabular}{r@{}l|c|cc|ccccccccccc}
			\toprule 
			\multicolumn{2}{c|}{Method} &  Mod.& mAP$\uparrow$ &     NDS $\uparrow$& Car  & Truck   & Bus   & Trailer  & C.V. & Ped. & Mot. & Byc. & T.C. & Bar.  \\
			\midrule
			
			
			Focals Conv~\cite{chen2022focal}&~\textcolor[HTML]{708090}{\scriptsize{[NeurIPS 22]}}& L&63.8&70.0  &86.7 &56.3& 67.7& 59.5& 23.8& 87.5& 64.5& 36.3& 81.4& 74.1        \\
			
			TransFusion-L~\cite{bai2022transfusion}&~\textcolor[HTML]{708090}{\scriptsize{[CVPR 22]}} &L   & 65.5       &   70.2   & 86.2 & 56.7 & 66.3& 58.8& 28.2& 86.1& 68.3& 44.2& 82.0& 78.2    \\
			
			Link~\cite{lu2023link}&~\textcolor[HTML]{708090}{\scriptsize{[CVPR 23]}} &L & 66.3    &   71.0   & 86.1 & 55.7 & 65.7& 62.1& 30.9& 85.8& 73.5& 47.5& 80.4& 75.5    \\
			LiDARMultiNet~\cite{ye2023lidarmultinet}&~\textcolor[HTML]{708090}{\scriptsize{[AAAI 23]}}    &L & 67.0       &   71.6   &86.9 &57.4 & 64.7& 61.0& 31.5& 87.2& 75.3& 47.6& \textbf{85.1}& 73.5   \\
			
			FSTR-L ~\cite{zhang2023fully}&~\textcolor[HTML]{708090}{\scriptsize{[TGRS 23]}}  &L & 67.2       &   71.5   & 86.5 & 54.1 & 66.4& 58.4& 33.4&88.6& 73.7& 48.1& 81.4& 78.1    \\
			HEDNet ~\cite{zhang2024hednet}&~\textcolor[HTML]{708090}{\scriptsize{[NeurIPS 23]}} &L &  67.7      &   72.0   &87.1 &56.5 &70.4&63.5& 33.6&  87.9& 70.4&44.8& 85.1& 78.1    \\
			
			FocalFormer3D~\cite{chen2023focalformer3d}&~\textcolor[HTML]{708090}{\scriptsize{[ICCV 23]}} &L & 68.7 & 72.6 &	87.2
			&57.1&69.6& 64.9&34.4 &88.2&\textbf{76.2} & 49.6&82.3 &77.8  \\
			SAFDNet~\cite{zhang2024safdnet}&~\textcolor[HTML]{708090}{\scriptsize{[CVPR 24]}} &L  & 68.3       &   72.3   &87.3 & 57.3 & 68.0& 63.7& 37.3&\textbf{89.0}& 71.1& 44.8&84.9& 	\textbf{79.5}    \\
			
			Co-Fix3D (Ours) & &L& \textbf{69.4} &\textbf{ 73.5 }&	\textbf{88.7}
			&\textbf{59.1}& \textbf{72.5}&  \textbf{65.5} &34.8 &87.2&71.8&\textbf{52.6}&83.7 &78.1  \\
			\midrule
			TransFusion~\cite{bai2022transfusion}&~\textcolor[HTML]{708090}{\scriptsize{[CVPR 22]}} &LC   &68.9      &  71.7   &87.1 & 60.0 &68.3& 60.8& 33.1& 88.4& 73.6& 52.9& 86.7& 78.1   \\	
			BEVFusion~\cite{liang2022bevfusion}&~\textcolor[HTML]{708090}{\scriptsize{[NerIPS 22]}} &LC &69.2 & 71.8 &88.1
			&60.9&69.3&62.1 &34.4 &89.2&72.2 & 52.2&85.2 &78.2  \\
			
			BEVFusion-MIT~\cite{liu2023bevfusion}&~\textcolor[HTML]{708090}{\scriptsize{[ICRA 23]}} &LC &70.2 & 72.9 &88.6
			&60.1&69.8&63.8 &39.3 &89.2&74.1 & 51.0&86.5 &80.0  \\
			DeepInteraction~\cite{yang2022deepinteraction}&~\textcolor[HTML]{708090}{\scriptsize{[NerIPS 22]}} &LC &70.8 & 73.4 &87.9
			&60.2&70.8&63.8 &37.5 &\textbf{91.7} &75.4 & 54.5&87.2 &\textbf{80.4}  \\
			
			ObjectFusion~\cite{cai2023objectfusion}&~\textcolor[HTML]{708090}{\scriptsize{[ICCV 23]}} &LC &71.0 & 73.3 &89.4
			&59.0&71.8&63.1 &40.5 &90.7 &78.1 & 53.2&87.7 &76.6  \\
			
			FocalFormer3D~\cite{chen2023focalformer3d}&~\textcolor[HTML]{708090}{\scriptsize{[ICCV 23]}}  &LC & 71.6 & 73.9 &	88.5
			&\textbf{61.4}&71.7 &\textbf{66.4 }&35.9 &89.7&\textbf{80.3} & 57.1&85.3 &79.3  \\
			GraphBEV~\cite{yan2023cross}&~\textcolor[HTML]{708090}{\scriptsize{[ECCV 24]}}   &LC     & {71.7}          & 73.6              &  89.2             & 60.0              &   72.1           &      64.5     &  \textbf{ 40.8  }          &    90.9    	&   76.8       &    53.3          & \textbf{ 88.9}       &       80.1      \\					
			
			Co-Fix3D (Ours)  & &LC& \textbf{72.3} & \textbf{74.7} &	\textbf{89.6}
			&60.8& \textbf{73.3}& 65.2 &37.9 &88.7&77.6 & \textbf{62.7}&86.7 &	80.0  \\
			\bottomrule
		\end{tabular}%
			\label{tab:nus_test}
	\end{center}		
	\vspace*{-3mm}	

\end{table*}	
		
		\begin{enumerate}

			\item Variant A (Fig. \ref{design_lge}.(a)) utilizes the $Wavelet\ Encode$ method to examine the ability of DWT to repair defective regions within BEV features.
			
			\item Variant B (Fig. \ref{design_lge}.(b)) employs the $Hybrid\ Encode$ approach to investigate the capability of the attention mechanism to repair defective regions within BEV features.
			
			\item B $\rightarrow$ C(Fig.\ref{design_lge}.(c)): Building on Variant B, $Wavelet\ Decode$ is added to enhance the output of the attention mechanism, further exploring the repair effectiveness in this context.
			\item C $\rightarrow$ D(Fig.\ref{design_lge}.(d)): 
			Building on Variant C, $Wavelet\ Encode$ is added to explore the repair effectiveness of DWT in this context. 
			
			\item D $\rightarrow$ E: Building on D, this variant E (Fig.\ref{design_lge}.(e)) starting with local pixel-level feature optimization $Wavelet\ Encode$, followed by $Hybrid\  Encode$, and concluding with $Wavelet\ Decode$.  
			\item D $\rightarrow$ F: Building on D, this variant F (Fig.\ref{design_lge}.(f)) starting with local pixel-level feature optimization $Wavelet\ Encode$, followed by $Wavelet\ Decode$ , and concluding with the $Hybrid\  Encode$.
			
			\item A $\rightarrow$ G: This variant G (Fig.\ref{design_lge}.(g)) performs global attention and local optimization in parallel, concatenating the results. This parallel processing strategy leverages the strengths of both global and local optimizations, aiming to achieve a more comprehensive and refined features effect through the combined outcomes.
		\end{enumerate}
		Each variant explores the best way to integrate global and local optimizations through different sequences and methods, aiming to achieve an optimal balance between detail restoration, noise suppression, and computational efficiency.


		\section{Experiments   }
		
		\subsection{Datasets }	
		The nuScenes Dataset~\cite{caesar2020nuscenes} is an extensive outdoor dataset, ideal for both LiDAR-based and multimodal LiDAR-camera integrated 3D object detection. It features 1,000 multi-modal scenes, each lasting 20 seconds, annotated at 2Hz.  This dataset includes data from a 32-beam LiDAR at a 20FPS rate and images from a 6-view camera setup. We evaluated our method under both LiDAR-only and LiDAR-Camera fusion settings, using the official nuScenes metrics: mean average precision (mAP) and nuScenes detection score (NDS). Our training and evaluation adhered to the nuScenes standard protocol, analyzing data from the preceding nine frames for current frame assessment, in line with the official evaluation criteria.
		
		\subsection{Implementation Details}	
		\label{implementation}
		We developed our model using the PyTorch and the open-source MMDetection3D \cite{contributors2020mmdetection3d}. The detection region spans $[-54.0m, 54.0m]$ on the X and Y axes, and $[-5.0m, 3.0m]$ on the Z axis. On the nuScenes dataset, we set the voxel size to 0.075m $\times$ 0.075m $\times$  0.2m.
		In LiDAR mode, the backbone was initially trained for 20 epochs using CBGS \cite{zhu2019class}. Subsequently, we froze the pre-trained LiDAR backbone and continued training the detection head with multi-stage heatmaps for an additional six epochs, employing GT sample augmentation except in the final five epochs.
		In multi-modality mode, The image backbone network utilizes ResNet-50 and  the image size set to 448 $\times$ 800, following the FocalFormer3D and BEVFusion approach, to project multi-view camera features onto a predefined voxel grid in 3D space. The BEV size of this voxel grid is set to 180 $\times$ 180, matching the 8 $\times$  downsampled BEV features generated by VoxelNet\cite{zhou2018voxelnet}, with a channel dimension of 128. The camera backbone was trained for 20 epochs without CBGS. Then both the image and point cloud branches were frozen, only the fusion module and head module gradients were enabled, and the training continued for 10 epochs without CBGS.
		Our model is trained with the total batch size of 16 on 4 Nvidia 4090 GPUs.  We utilize the AdamW\cite{loshchilov2017decoupled} optimizer for the optimization process. The initial learning rate is set to $1.0\times 10^{-4}$, and we apply the one-cycle policy for learning rate adjustment.
		
		\subsection{ State-of-the-Art Comparison}
		
		\paragraph{LiDAR-Based 3D object detection on test set} 			
		In Tab.~\ref{tab:nus_test}, we benchmarked the performance of our model on the nuScenes test set and compared it with the current leading LiDAR-based ('L') and multimodal ('LC') 3D object detectors. The results demonstrate that Co-Fix3D outperforms all existing state-of-the-art (SOTA) 3D detection algorithms. As a baseline for TransFusion-L, Co-Fix3D's LiDAR mode achieved a 3.9\% improvement in mAP and a 3.3\% increase in NDS. Additionally, compared to recent single-modal detection methods such as HEDNet, SAFDNet, and FocalFormer3D, Co-Fix3D exhibited superior performance, with mAP gains of 1.7\%, 1.1\%, and 0.7\%, respectively. Notably, Co-Fix3D achieved the highest detection results in certain categories, such as car, truck, and trailer. This suggests that Co-Fix3D effectively enhances BEV features through parallel LGE, enabling more accurate identification of weak positive queries.
		
		\paragraph{Multi-modal 3D object detection on test set}
		We extended Co-Fix3D as a multimodal model and used it as a baseline for TransFusion-LC. In its multimodal mode, Co-Fix3D improved mAP by 3.4\% and NDS by 3.0\%. Furthermore, compared to the latest single-modal detection methods such as ObjectFusion, GraphBEV, and FocalFormer3D, Co-Fix3D exhibited superior performance, with mAP gains of 1.3\%, 0.6\%, and 0.7\%, respectively. Notably, Co-Fix3D achieved the highest detection results in certain categories, such as car and bus. This further demonstrates that Co-Fix3D enhances BEV features through parallel LGE, effectively improving overall detection capability.
		
		\begin{table}[h]
	\renewcommand{\arraystretch}{1.1}
	
	\begin{center}
			\caption{
			Comparison with detectors on the nuScenes \textbf{VALIDATION} set.	Mod.: Modality.	}
		\begin{tabular}{l|c|c|cc}
			\toprule 
			Method & \small{Mod.}  & \begin{tabular}[c]{@{}l@{}}Image\\ Encoder\end{tabular} & mAP  & NDS  \\	
			\midrule
			TransFusion-L~\cite{bai2022transfusion} &L  & &64.9& 69.9  \\
			HEDNet ~\cite{zhang2024hednet} &L  & & 66.7 & 71.4 \\							
			
			SAFDNet~\cite{zhang2024safdnet}&L   & & 66.3 & 71.0 \\
			FocalFormer3D~\cite{chen2023focalformer3d}&L  & & 66.5 &71.1 	\\
			\textbf{Co-Fix3D (Ours)}  &L   & & \textbf{67.3} & \textbf{72.0}  	\ \\
			\midrule
			TransFusion~\cite{bai2022transfusion}  &LC  &\small{ResNet-50} &67.5&  71.3 \\
			BEVFusion~\cite{liu2023bevfusion} & LC  &\small{Swin-T} &68.5 & 71.4 	\\
			SparseFusion\cite{xie2023sparsefusion}  &LC &\small{ResNet-50}& 70.4 & 72.8	\\		
			FocalFormer3D~\cite{chen2023focalformer3d}& LC  &\small{ResNet-50}  & 70.5 & 73.0 	\\	
			\textbf{Co-Fix3D (Ours)}&LC  &\small{ResNet-50}& \textbf{70.8} & \textbf{73.6}	\\
			\bottomrule
		\end{tabular}
		\label{tab:nus_val}	
	\end{center}	

	\vspace*{-2mm}
\end{table}
		
		\paragraph{3D object detection on val set} 	
		We present results on the nuScenes validation set, as detailed in Table \ref{tab:nus_val}. As a baseline for TransFusion-L, Co-Fix3D's LiDAR mode achieved a 2.4\% improvement in mAP and a 2.1\% increase in NDS. Additionally,  Co-Fix3D enhances the LiDAR-only baseline, FocalFormer3D, with an increase of 0.8\% in mAP and 0.9\% in NDS. For multi-mode scenarios, the improvement is 0.3\% in mAP and 0.6\% in NDS.

		\subsection{Ablation Study} 	
		We conducted several experiments on the validation set. We conducted 20 epochs of training tests, implementing a degradation strategy in the last five epochs.
		
		\begin{table}	
	\renewcommand\arraystretch{1.1}	
		\caption{{\textbf{Ablation studies for each Stage in Co-Fix3D} on the nuScenes validation set. Baseline-L indicates the LiDAR-only baseline, which refers to a simple variant of  Co-Fix3D without employing the LGE or Parallel Structure(P.S.).}}	
	\begin{tabular}{l|ccccc|ll}
		\toprule
		&	Baseline-L  & LGE& P.S.&Stages &Queries  &mAP$\uparrow$ & NDS$\uparrow$\\	
		\midrule		
		(a) &	 \checkmark& & &1&200&65.2 &70.4 \\
		(b)&&\checkmark& &1&200&66.3 &71.6 \\	
		\midrule	
		(c) & \checkmark&&&2&400&66.3 &70.8 \\	
		(d) && \checkmark&&2&400&66.8 &71.6 \\
		(e) && \checkmark&\checkmark&2&400&67.0 &71.7\\
		(f) && \checkmark&\checkmark&2&600&67.1 &71.8\\
		\midrule
		(g)	&\checkmark&& &3&600&66.5 &71.1 \\	
		(h)	&&\checkmark& &3&600&67.1 &71.9\\	
		(i)	&&\checkmark&\checkmark &3&600&67.3 &72.0\\				
		\bottomrule
	\end{tabular}

	\label{tab:lgefuc}
\end{table}			
		
		\paragraph{Advantages of LGE} Tab.\ref{tab:lgefuc} primarily illustrates the advantages of the LGE module. For instance, in the first stage, incorporating the LGE structure improved mAP by 1.1\% and NDS by 1.2\%, significantly enhancing detection performance. In subsequent stages, configurations with the LGE module consistently outperformed those without it. Even within the cascade structure, setups with LGE outperformed those lacking the module. Furthermore, when comparing cascade and parallel structures, the parallel structure with LGE demonstrated superior performance. At stage 3, using the parallel LGE structure resulted in mAP and NDS improvements of 0.8\% and 0.9\%, respectively. These findings suggest that the LGE module effectively refines BEV features and filters out weak positive queries, thereby improving the recognition rate of hard-to-detect samples in later stages and enhancing overall mAP.
		
		\begin{figure}[h]
			\centering
			\includegraphics[width=1.0\linewidth]{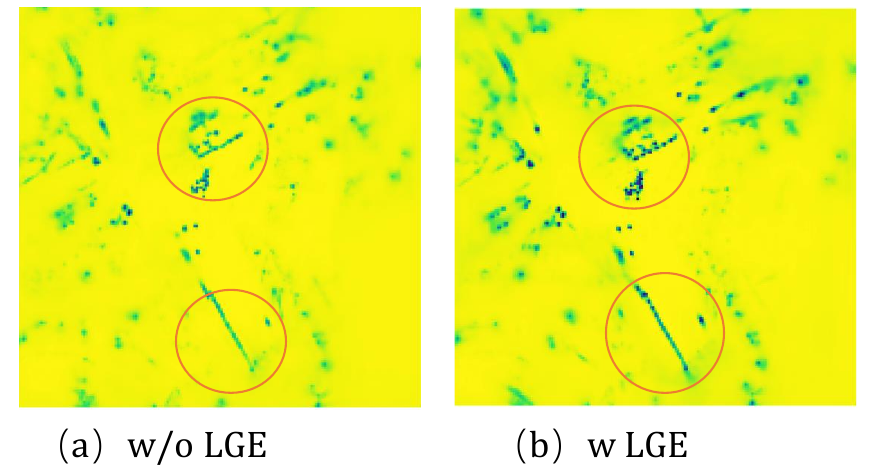}	
			\caption{The impact of LGE on features. By comparing (a) and (b), we found that the features within the red area in (b) are significantly better than those in (a).}
			\label{feature}
		\end{figure}
		
		\begin{figure*}[t]
			\centering
			\includegraphics[width=0.91\linewidth]{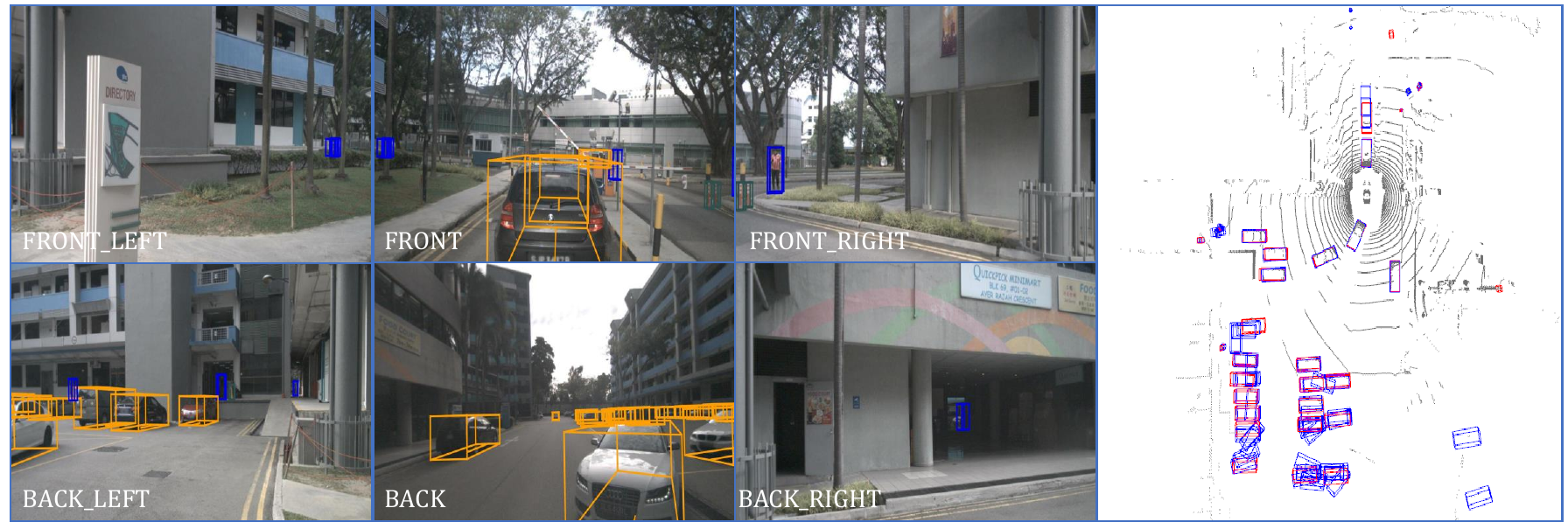}	
			\caption{Examples of 3D object detections on the nuScenes validation set. In the rightmost point cloud image, the red boxes represent the GTs, and the blue boxes denote the predictions. The total number of boxes displayed is set at 100.}
			\label{viz}
		\end{figure*}	
			\paragraph{Feature Visualization} To investigate the impact of the LGE module on BEV feature maps, we conducted a visualization analysis of the BEV features, as shown in Fig.\ref{feature}. The images reveal that the BEV features processed with the LGE module are significantly better than those without it, indicating that the LGE module can repair some defective features, thereby effectively enhancing the quality of queries in TransFusion-type models.

		\begin{table}[h]	
			\renewcommand\arraystretch{1.1}
						\caption{Ablation studies on the impact of increasing the number of queries during the testing phase on the results.}	
			\begin{tabular}{l|cccc|ll}
				\toprule
				&	Baseline-L  & LGE& P.S. &Queries  &mAP$\uparrow$ & NDS$\uparrow$\\	
				\midrule	
				(1) &	 \checkmark&&& 600&66.5 &71.1 \\
				(2) &	 \checkmark&&&  900&66.5 &71.1 \\
				(3) &	 \checkmark&&&  1200&66.5 &71.1 \\
				\midrule	
				(4) &	 &\checkmark&\checkmark&600&67.3 &72.0 \\
				(5) &	 &\checkmark&\checkmark&900&67.4 &72.1 \\			 	
				(6) &	 &\checkmark&\checkmark&1200&67.5 &72.2\\	
					
				\bottomrule
			\end{tabular}

			\label{tab:test_lge}
		\end{table}
		
		\paragraph{Cascaded Structure vs. Parallel Structure} We compared parallel and cascaded structures, both using LGE modules but configured differently. While maintaining 300 bounding boxes, we increased the query count to 1200 as shown in Table \ref{tab:test_lge}. The parallel structure achieved a 0.2\% improvement in both mAP and NDS over the cascaded structure. In the cascaded setup, raising the query count from 600 to 1200 yielded no significant improvement, likely due to the first-stage LGE's modifications to the BEV features, which may have limited further enhancements by later modules. In contrast, the increased queries in the parallel structure demonstrated its greater effectiveness in enhancing detection performance.

	\begin{table}[h]	
	\centering	
	\caption{Experiments on Different Variants of LGE}
	\setlength{\tabcolsep}{5.9mm}{}	
	\renewcommand\arraystretch{1.00}	
	\begin{tabular}{cccc}	
		\toprule
		Variant  & mAP$\uparrow$ & NDS$\uparrow$ \\	
		\midrule
		(A$_0$) &65.2 &70.4 \\	
		(A) &65.9 &71.1 \\
		(B) &65.5 &70.6 \\
		(C) &65.6 &70.7 \\
		(D) &65.4 &70.5 \\
		(E) &N/A &N/A \\
		(F) &65.8 &71.0 \\
		(G) &66.3 &71.6 \\
		\bottomrule
	\end{tabular}

	\label{tab:lgedegin}
\end{table}
	
	\paragraph{Design choices of LGE}
To more effectively assess the effectiveness of the LGE module, we conducted detailed experiments on various LGE design choices, with results shown in Table \ref{tab:lgedegin}. Variant A$_0$ serves as the baseline performance without any enhancement module. Variant A uses local optimization with DWT, resulting in increases of 0.7\% in mAP and 0.7\% in NDS. Variant B, which solely utilizes a global attention mechanism, shows limited improvements of 0.3\% in mAP and 0.2\% in NDS, indicating that global feature enhancement alone offers minimal performance gains. Variant C adds a Wavelet Decode after global attention optimization, achieving only a 0.1\% improvement over Variant B. Variant D, building on C, first applies global attention followed by the DWT module, which unfortunately leads to a decrease in performance. Variant E experiences training anomalies such as gradient issues and non-convergence due to improper configuration. Variant F, which builds upon Variant A by adding global attention, results in a 0.1\% decrease in mAP, suggesting that additional global optimization may be detrimental when local features are already well-optimized. Ultimately, Variant G, which implements simultaneous local and global optimizations, shows our proposed method, enhancing mAP and NDS by 1.1\% and 1.2\%, respectively. Therefore, we have chosen Variant G as our LGE module.

	\paragraph{Visualization} Fig. \ref{viz} displays our qualitative results on the nuScenes validation set. It can be seen that the performance of 3D object detection is quite good.
		\begin{table}[h]	
		\centering	
		\caption{Experiments focusing on one of the hyperparameters for the LEG module, specifically the number of iterations for DWT.}
		\setlength{\tabcolsep}{5.9mm}{}	
		\renewcommand\arraystretch{1.00}	
		\begin{tabular}{cccc}	
			\toprule
			num  & mAP$\uparrow$ & NDS$\uparrow$ \\	
			\midrule
			1 &65.8 &70.9 \\	
			2 &66.1 &71.3 \\
			4 &66.3 &71.6 \\
			6 &66.2 &71.5 \\
			\bottomrule
		\end{tabular}	
		
		\label{tab:lgecount}
	\end{table}
	\paragraph{Ablation studies for LEG} Table \ref{tab:lgecount} shows the impact of the number of iterations on the results in the LEG module, where DWT and attention mechanisms are applied in parallel.
	Table \ref{tab:latency} presents the latency and parameter count.

	\begin{table}[h]
		\begin{center}
				\caption{Latency analysis for model components. Latency is measured on a 4090 GPU for reference. }
				\begin{tabular}{lccccc}
					\toprule
					Models/Components & Latency  &Parameters\\ \midrule
					TransFusion-L & 78ms &18.24M\\ \midrule
					FocalFormer3D-L & 94ms &22.63M \\ 
					\midrule
					Co-Fix3D-L & 123ms & 30.13M\\ 
					--   Multi-stage LGE  & 34ms  & 9.76M\\
					--   Other & 89ms   &20.17M\\
					
					\bottomrule
				\end{tabular}
						\label{tab:latency}
			\end{center}

		\end{table}	
		
		\section{Conclusion}
		We introduced Co-Fix3D, an end-to-end 3D object detection network designed to enhance BEV features and leverage a collaborative network approach for comprehensive mining of potential samples. This method significantly boosted the performance of 3D object detection in autonomous driving scenarios. Extensive  experiments confirmed that Co-Fix3D not only excels in single-modality point cloud detection but also in hybrid point cloud-image fusion modalities, achieving state-of-the-art performance on the nuScenes benchmark.  We believe that Co-Fix3D will serve as a robust and efficient baseline for future research in this field.
{	
	\small
	\bibliographystyle{IEEEtran}
	\bibliography{root}
}
\end{document}